\documentclass[a4paper,fleqn]{cas-dc}
\usepackage{algorithm, algorithmic}
\usepackage[numbers,sort&compress]{natbib}
\usepackage{ulem}
\usepackage{cleveref}
\usepackage{color}
\usepackage{subfigure}
\usepackage{diagbox}

\usepackage[font=small,labelfont=bf,labelsep=none]{caption}
\usepackage{caption}
\usepackage{stfloats}
\begin{document}
\fontsize{9pt}{10pt}\selectfont
\let\WriteBookmarks\relax
\def\floatpagepagefraction{1}
\def\textpagefraction{.001}

\shorttitle{Enhancing robustness of data-driven SHM models: adversarial training with circle loss}

\shortauthors{Yang et~al.}

\title [mode = title]{Enhancing robustness of data-driven SHM models: adversarial training with circle loss} 



%
\author[1]{Xiangli Yang}
\ead{xiangli.yang@cqjtu.edu.cn}
\credit{Conceptualization of this study, Methodology, Software}

\author[1]{Xijie Deng}
\ead{xijie.deng@mails.cqjtu.edu.cn}
\credit{Conceptualization of this study, Methodology, Software}
\cormark[1]

\author[2,3]{Hanwei Zhang}
\ead{zhang@depend.uni-saarland.de}
\credit{Conceptualization of this study, Methodology, Software}

\author[4]{Yang Zou}
\ead{yang.zou@auckland.ac.nz}
\credit{Conceptualization of this study, Methodology, Software}

\author[1]{Jianxi Yang}
\ead{yjx@cqjtu.edu.cn}
\credit{Conceptualization of this study, Methodology, Software}

 \affiliation[1]{organization={School of Information Science and Engineering, Chongqing Jiaotong University},
     city={Chongqing},
     postcode={400074}, 
     country={China}}
\affiliation[2]{organization={Institute of Intelligent Software},
    city={Guangzhou},
    postcode={511400},
    country={China}}
\affiliation[3]{organization={Saarland University},
    city={Saarbrücken},
    postcode={66123},
    country={Germany}}
\affiliation[4]{organization={Department of Civil and Environmental Engineering, University of Auckland},
    city={Auckland},
    postcode={1023},
    country={New Zealand}}

\cortext[cor1]{Corresponding author}

\begin{abstract}
Structural health monitoring (SHM) is critical to safeguarding the safety and reliability of aerospace, civil, and mechanical infrastructure. 
Machine learning-based data-driven approaches have gained popularity in SHM due to advancements in sensors and computational power. However, machine learning models used in SHM are vulnerable to adversarial examples \textemdash even small changes in input can lead to different model outputs. This paper aims to address this problem by discussing adversarial defenses in SHM. In this paper, we propose an adversarial training method for defense, which uses circle loss to optimize the distance between features in training to keep examples away from the decision boundary. Through this simple yet effective constraint, our method demonstrates substantial improvements in model robustness, surpassing existing defense mechanisms.
\end{abstract}


\maketitle

\section{Introduction}
Structural Health Monitoring (SHM) serves to continuously monitor and evaluate the condition of engineering structures in real-time, alerting in advance when anomalies in the structure's health arise, and offering tailored solutions for various issues. The SHM process involves the installation of sensors in the engineering structure to collect response data within its environment. Subsequently, a set of indirect tools is utilized to analyze this data, aiming to detect, pinpoint, and manage any structural damage present. The existing methods for SHM can be categorized into two main types.
The first type, known as the \textit{model-driven method}, aims to create a precise finite element model by adjusting model parameters using input-output measurements. This method involves comparing these model parameters with actual structural measurement data to identify structural conditions. However, it is highly dependent on the accuracy of the theoretical model and the quality of the monitoring data. Additionally, it requires personnel with specialized knowledge in bridge modeling. Yet, these models are often static and might not accommodate unforeseen changes or new parameters, necessitating a ‘re-modeling' process when such alterations occur.
On the other hand, the second type, known as the \textit{data-driven method}, constructs a model in the form of statistical representations. This approach detects changes in the structural state by analyzing evolving patterns and probability distributions within the monitoring data itself.

Data-driven methods in SHM are gaining popularity due to recent technological advancements in sensors, high-speed internet, and cloud-based computation. The data-driven method is based on the machine learning (ML) paradigm, which allows for the rapid and inexpensive creation of effective models that are widely used for structural diagnosis and structural damage detection. Some pioneering studies of ML methods have been conducted in data-driven SHM, including the use of Bayesian networks~\cite{Bayesianbook, arangio2015structural, YIN2017260}, artificial neural networks (ANN)~\cite{yang2020hierarchical, yang2021data, he2022framework}, and support vector machines~\cite{WIDODO20072560, prasanna2014automated, kim2019crack, Yang2021bridge}. However, several studies have shown that ML models are vulnerable to adversarial examples~\cite{szegedy2013intriguing, goodfellow2014explaining, biggio2013evasion, nguyen2015deep, zhang2022deep}, that is, samples with perturbations designed to deceive the ML model and cause it to predict an incorrect label with high confidence. Adversarial perturbation is usually imperceptible to humans and difficult to detect from the original sample.

The expanding use of ML models in SHM has raised concerns due to opaque decision-making processes and a lack of interpretability in certain models, especially in critical fields where safety is paramount. Furthermore, the emergence of adversarial examples has heightened these concerns.
Max et al.~\cite{champneys2021vulnerability} have demonstrated the serious vulnerability of data-driven SHM models to adversarial attacks concisely and efficiently. 
Consequently, ensuring robustness against adversarial attacks has become a key challenge in the broad implementation of SHM frameworks. Adversarial robustness, which signifies a model's capability to withstand adversarial examples, remains an area largely unexplored within the SHM field.

In computer vision, there exist two approaches for bolstering adversarial robustness: \textit{proactive defenses}~\cite{xie2019feature,buckman2018thermometer,zhang2021patch}, which involve altering or identifying input data during the inference phase, and \textit{active defense}~\cite{oberman2018lipschitz,li2021ensemble,wong2020fast}, which entails adjusting the fundamental framework or learning process during the training phase.
Transformations, a typical reactive approach~\cite{xu2017feature,guo2017countering,rebuffi2021data}, aim to neutralize adversarial effects through the application of simple filters. Although cost-effective, this method fares poorly against potent attacks like PGD~\cite{madry2017towards}, C\&W~\cite{carlini2017towards}, and DeepFool~\cite{moosavi2016deepfool}. To enhance its efficacy, transformations introduce randomness~\cite{raff2019barrage,prakash2018deflecting} and representation~\cite{moosavi2018divide,buckman2018thermometer,liu2019feature}. While this augmentation boosts robustness, it often compromises accuracy by altering original images, thereby discarding adversarial context. Consequently, networks trained on original data may struggle to recognize distorted information.
Adversarial Training~\cite{goodfellow2014explaining,madry2017towards,li2023data}, a widely-used proactive defense strategy, enriches the training process by incorporating adversarial images. This approach enables the network to learn from adversarial instances, enhancing its comprehension of relevant knowledge. Adversarial training employs specific attacks to generate adversarial images. However, as various attacks possess different characteristics, the model becomes susceptible to unseen attack methods.

To boost adversarial training, we introduce a novel adversarial training loss, namely circle loss~\cite{sun2020circle}, which enhances adversarial robustness by limiting the distance between samples within the feature space. By promoting better compactness within classes and minimizing discrepancies between classes, the circle loss indirectly shifts data points away from the decision boundary. This not only enhances robustness but also prevents the model from overfitting to the specific adversarial examples used during training.
Furthermore, we use the circle loss as a regularization term, integrating it into the standard training procedure. The experimental results show that this approach aims to enhance the overall robustness of the training models.

The main contributions of this paper can be summarized as follows: 
\begin{enumerate}[\hspace{2em}1.]
    \item We delve into the adversarial phenomenon within SHM, conducting a comprehensive analysis of the threats and establishing an adversarial attack threat model specifically tailored for the SHM field.
    \item We analyze the adaptability and robustness of applying existing defense methods in SHM and explore potential directions for adversarial machine learning within the SHM domain.
    \item We introduce an adversarial training methodology for defense, optimizing the feature distances during training to ensure examples remain distant from the decision boundary. This method significantly improves the adversarial robustness of data-driven SHM models.
\end{enumerate}

The rest of the paper is organized as follows. ~\autoref{S2} presents the adversarial threats in data-driven SHM. ~\autoref{S3} introduces our defense method. ~\autoref{S4} demonstrates the performance of our defense method. Conclusions are drawn in ~\autoref{S5}.

\section{Adversarial vulnerability of data-driven SHM}
\label{S2}

In this section, we first define adversarial vulnerability and the scope of threat models in SHM. Then we introduce the existing adversarial attacks and defense strategies.

\subsection{Threat model}
Given a trained model $\mathcal{F}$, an original input data sample $x$, generating an adversarial example $x^{\mathrm{adv}} $ can generally be described as an optimization problem:
\begin{equation}\label{eq:adv}
    \begin{aligned}
         \operatorname{minimize}  & ~\mathcal{D}(x, x^{\mathrm{adv}}) \\
        \text { such that } & \mathcal{F}(x)\ne \mathcal{F} (x^{\mathrm{adv}})
    \end{aligned}
\end{equation}
where $\mathcal{D}$ is some distance metric. 
By minimizing the difference between $x^{\mathrm{adv}}$ and $x$ with the constrain, we force the modification is imperceptible
(i.e., $x \approx x^{\mathrm{adv}}$ ) while the model output is completely different. 

\paragraph{Adversary's knowledge.} According to the adversary's knowledge we have two different settings, i.e., attacks under white-box settings and black-box settings.
White-box attacks assume the adversary possesses complete knowledge about the targeted model, including its structure, parameters, and training data. 
In contrast, black-box attacks assume that the adversary is only able to observe its output (labels or confidence scores), which is more realistic and aligns better with real-world threat models. White-box attacks are crucial since adversarial examples' transferability allows white-box attacks to corrupt ML models in black-box settings by transferring the adversarial effects from surrogate models to target models~\cite{papernot2016transferability, zhao2022towards}.

\paragraph{Adversary's goal.} Adversarial goals in altering the classifier output are broadly classified as untargeted and targeted attacks. Untargeted attacks aim to change the classification output to any different class from the original, essentially misrepresenting the structural state by minimizing the likelihood of the correct class. While targeted attacks aim to force the output classification into a specific target class. For instance, causing a mislabeling of a damaged state as undamaged (false negative) by maximizing the probability of the desired target label. Targeted attacks require larger perturbations to succeed due to limited space available to redirect examples toward a specific label~\cite{liu2017delving}.

\paragraph{Adversary's capability.} In addition to deceiving the model, the adversarial examples are supposed to stay stealthy. In computer vision, most work constrains adversarial perturbations by using $L_p$ norm or perceptual metrics~\cite{zhang2020smooth} as $\mathcal{D}$.
Different from computer vision, SHM models receive signals from sensors to evaluate the structure's state. Such signals are original with small noises barely recognizable by humans. 
Thus the distance metric $\mathcal{D}$ for attacking SHM models is supposed to focus on preserving damage-sensitive information in original samples, which differs in two scenarios:
(1) in cases where engineers quickly estimate time-domain features from sensor data visually, $\mathcal{D}$ should preserve these features related to structural damage;
(2) in cases where sensor data requires further signal processing and feature extraction, e.g., calculating Frequency Response Functions (FRFs) for intuitive diagnosis, $\mathcal{D}$ should focus on preserving damage-related information, such as FRF peaks and valleys.
The choice of the budget of distortion should consider the normal noise level in SHM to ensure stealthiness. Inherent noise, stemming from environmental sources and system components, increases engineers' tolerance for noise and offers opportunities for adversarial perturbations.

\subsection{Adversarial attacks}
\label{S2.2}
Existing image adversarial example generation methods can be categorized as gradient-based, optimization-based, and generator-based, which provide valuable references for attacking SHM models.
\textit{Gradient-based methods} generate adversarial examples by perturbing along the gradient direction of a differentiable model. FGSM~\cite{goodfellow2014explaining}, as a first try, searches for adversarial examples along the sign of gradients of the loss function with a given step size once.
Techniques like Basic Iterative Method (BIM)~\cite{kurakin2018adversarial} extend FGSM by applying it iteratively.
In black-box settings, Zeroth Order Optimization (ZOO)~\cite{chen2017zoo} and other gradient-based methods like~\cite{ilyas2018prior, ilyas2018black, tu2019autozoom, bhagoji2017exploring} estimate the gradients directly to generate adversarial examples.
\textit{Optimization-based methods} frame adversarial sample generation as an optimization problem, like the C\&W method~\cite{carlini2017towards}. The C\&W method customizes different objective functions $f$, users can choose the optimal objective function through experiments to achieve adversarial attacks. 
\textit{Generator-based methods} like AdvGAN~\cite{xiao2018generating} use parameterized generative adversarial networks (GAN) to create attacks, where GANs' generators and discriminators compete to generate high-quality adversarial examples.

\subsection{Defense strategies}\label{S2.3}

Within computer vision, there are two strategies to enhance adversarial robustness: proactive defenses, which alter or identify input data during inference, and active defense, which modifies the core framework or learning process during training. To further classify existing approaches, proactive defense comprises \textit{data preprocessing} and \textit{adversarial detection}, while active defense includes \textit{gradient masking} and \textit{adversarial training}.
Data preprocessing involves techniques that effectively compress and transform input data to mitigate the impact of adversarial noise~\cite{meng2017magnet, xie2019feature, buckman2018thermometer, zhang2021patch, li2021ensemble}. Adversarial detection consists of methods that identify adversarial examples before feeding data into the model~\cite{metzen2017on, lu2017safetynet}. Gradient masking refers to techniques that obscure model gradients during inference to hinder the direct construction of adversarial examples against the model~\cite{papernot2016distillation, ba2014deep}. Adversarial training involves training the network using adversarial examples to fortify it against attacks~\cite{zhang2019theoretically, robey2021adversarial}.

Proactive defenses aim to boost adversarial resilience without changing the model, relying on transformations or detection techniques. However, this strategy often sacrifices accuracy to prioritize robustness. Additionally, the defense strategies commonly used in computer vision aren't easily adaptable to SHM models.
Several data preprocessing techniques predominantly adjust image-specific traits like bit depth and pixel values, which are unsuitable for SHM data. Similarly, methods solely altering frequency and vibration amplitude lack effectiveness in fortifying SHM systems against adversarial attacks. Moreover, employing an adversarial detection approach adds computational overhead, negatively impacting real-time performance.
Consequently, we prioritize active defenses that involve modifying the model or its learning process to strike a better balance between accuracy and robustness within SHM systems.

In the context of masking gradients, \textit{Randomized Smoothing} (RS)~\cite{cohen2019certified} introduces random noise around input data points, creating a smoothed model that mitigates the impact of small perturbations on predictions. Employing Monte Carlo methods allows for estimating prediction uncertainties and establishing a certified radius around data points, ensuring stable predictions within defined bounds and further strengthening the model's robustness.
On the contrary, \textit{Distillation}~\cite{papernot2016distillation} involves training a smaller model (student) by leveraging insights from a larger, more accurate model (teacher). In this Distillation Defense, both the student and teacher models share the same structure. By learning from the teacher's softer outputs (probabilities or logits), the student aims to replicate the teacher's behavior, enhancing its resilience against adversarial attacks. The gradient masking technique diminishes model accuracy and proves ineffective against attacks not reliant on gradients. Within this context, adversarial training stands out as the most appropriate defense for SHM systems.

\subsubsection{Adversarial training}
Adversarial training can be traced back to~\cite{goodfellow2014explaining}, in which models were strengthened by producing adversarial examples and injecting them into training data. 
Later, Shaham et al.~\cite{shaham2018understanding} proposed a formulation of adversarial training, which has been theoretically and empirically justified. Resembling a game between the attacker and the defender, adversarial training can be formulated as a minimax optimization problem as
\begin{equation}\label{eq:advtrain}
    \min_{\theta }  \sum_{i=1}^{m} \max_{\tilde{x}_i \in \mathcal{U}_i} \mathcal{L}(\theta ,\tilde{x}_i,y_i)
\end{equation}
in which $\mathcal{U}_i$ is the uncertainty set corresponding to sample $x_i$, $y_i$ denotes the label of $x_i$ and $\mathcal{L}(\cdot)$ is a classification loss (i.e., cross-entropy) . Madry et al.~\cite{madry2017towards} gives a reasonable interpretation of this formulation: the inner problem aims at generating adversarial examples by maximizing the training loss while the outer one guides the network in the direction that minimizes the loss to resist attacks. With such a connection, they use the adversarial examples generated by the Projected Gradient Descent (PGD) attack method as a solution for the inner problem. We refer to such an adversarial training as \textit{PGD-based adversarial training} (PGD-AT). Through extensive experiments, their approach significantly increases the adversarial robustness of deep learning models against a wide range of attacks, which is a milestone of adversarial training methods.
To improve efficiency,
\textit{Fast adversarial training} (Fast-AT)~\cite{wong2020fast} employs the FGSM method with random initialization to generate adversarial examples. It aims to efficiently improve the model's robustness against adversarial attacks by reducing the computational overhead of generating and incorporating adversarial examples during training.

\begin{figure*}[hpbt]
    \centering
    \includegraphics[width=16cm]{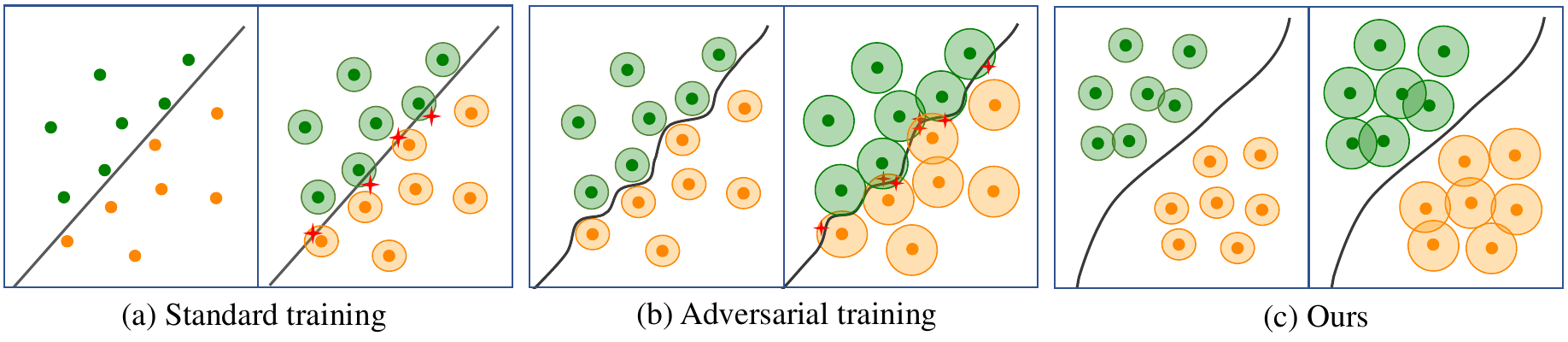}
    \caption{A conceptual illustration of decision boundaries after different training. Each circle represents a sample and its adversarial space within a perturbation budget $\epsilon$. In standard training, samples of different classes can be easily separated by a simple decision boundary, but this simple decision boundary cannot separate samples with adversarial perturbations. So some adversarial examples (noted by red stars) are misclassified. Conventional adversarial training learns a more complex decision boundary and separates samples with a certain perturbation budget, but it is powerless for samples with a larger perturbation budget. Our method can withstand large perturbations budget.}
    \label{fig:decisionboundary}
\end{figure*}

An assumption for adversarial phenomena is that data points reside near the model's decision boundary. Consequently, minor perturbations can push these points across the boundary, weakening the robustness of classification models. Adversarial training widens the gap between data points and the decision boundary, making it challenging for minor perturbations to shift data points to the opposite side, as depicted in ~\autoref{fig:decisionboundary}(a)(b). This enhancement accounts for the model's increased robustness after adversarial training. 
However, this also signifies that the effectiveness of adversarial training is constrained by the distortion levels of adversarial examples during training. By elevating the distortion budget, we can compromise the effectiveness of the adversarial training model in the inference phase.

To further strengthen adversarial training, we propose an enhancement integrating techniques from metric learning. Our aim is to optimize feature distances, providing a more effective defense against adversarial attacks.
To augment adversarial training, related researches~\cite{ross2018improving, zheng2016improving, moosavi2019robustness, kannan2018adversarial} impose regularization constraints on the discrepancy between the output of adversarial examples and their correct labels. Differing from them, our method employs regularization terms to ensure discrimination in the feature space between classes.
Adversarial training based on triplet loss~\cite{mao2019metric, li2019improving} applied metric learning methods to optimize feature distances between samples through similarity pairs, i.e., triples.
Our approach implements circle loss for adversarial training, an unexplored method, eliminating the need for constructing triples and pre-training a model beforehand.

\section{Methodology}\label{S3}

This section begins by presenting the concept of circle loss. Then we introduce how to apply circle loss with adversarial training.

\subsection{Circle loss}

Metric learning can be divided into two basic paradigms based on different optimization goals: Learning with Class-Level Labels and Learning with Pair-Wise Labels. Given the class label, the first method mainly learns to classify each training sample into the target category through the softmax classification loss function. Given a pair of labels, the second method directly learns the pairwise similarity in the feature space, generally using contrastive loss, triplet loss, etc.. Given a single sample $x$ in the feature space, assume that there are $K$ within-class samples and $L$ between-class samples associated with $x$, class-level label learning and pairwise label learning can be unified with the following loss:

\begin{equation}
\label{eq:uni}
      \begin{aligned}
        \mathcal{L}_{\mathrm{uni}}&=\log_{}{\left [ 1+\sum_{i=1}^{K}\sum_{j=1}^{L}\exp \left ( s_n^j-s_p^i+m \right )\right ] } \\
        &=\log_{}{\left [ 1+\sum_{j=1}^{L}\exp (\gamma (s_n^j+m))\sum_{i=1}^{K}\exp (\gamma (-s_p^i)\right ] }
      \end{aligned}
\end{equation}
in which ${\left \{s_p^1, s_p^2,\cdots, s_p^K \right \}}$ is the intra-class similarity score between $x$ and similar samples, $\left \{ s_n^1, s_n^2,\cdots,s_n^L \right \}$ is the similarity between $x$ and The inter-class similarity score between heterogeneous samples, $\gamma$ is the scaling factor. $m$ represents the margin, which represents the gap between the intra-class distance and the inter-class distance. It can be intuitively seen that this formula reduces $(s_n^j-s_p^i)$ by traversing each similar pair. In addition, with adjustments, this formula can also degenerate into a classification loss in class label learning or a triplet loss in pairwise label learning.

Circle loss was proposed by Sun et al.~\cite{sun2020circle} in the field of metric learning in 2020. It is based on the equation~\eqref{eq:uni} but can allow each similarity score to adjust its weight according to the current optimization status, with a high degree of optimization flexibility and clearer convergence. state. The formula for circle loss is as follows

\begin{equation}
\label{eq:circle}
      \begin{aligned}
        \mathcal{L}_{\mathrm{circle}}&=\log_{}{\left [ 1+\sum_{i=1}^{K}\sum_{j=1}^{L}\exp \left ( \alpha _n^j s_n^j - \alpha_p^i s_p^i \right )\right ] } \\
        &=\log_{}{\left [ 1+\sum_{j=1}^{L}\exp (\alpha _n^j s_n^j)\sum_{i=1}^{K}\exp (\gamma (-\alpha_p^i s_p^i)\right ] }
      \end{aligned}
\end{equation}
in which, $\alpha _n^j$ and $\alpha_p^i$ are non-negative weighting factors. It can be seen that compared with equation \eqref{eq:uni}, equation~\eqref{eq:circle} removes the margin $m$, and uses $\alpha _n^j s_n^j - \alpha_p^i s_p^i$ instead of $( s_n^j-s_p^i)$. During the training process, the gradient for$\alpha _n^j s_n^j - \alpha_p^i s_p^i$ will be multiplied by the weights $\alpha_n^j$ and $\alpha_p^i$ respectively when backpropagating to $s_n^j$ and $s_p^i$ . When a similarity score moves away from its optimal value, its weight $\alpha$ will increase, allowing it to be updated at a larger pace. The adjustment methods of $\alpha_n^j$ and $\alpha_p^i$ are defined as

\begin{equation}
      \begin{aligned}
        \alpha _n^j=\left [  s_j^n-O_n\right ] _+ \\
        \alpha _p^i=\left [  O_n-s_p^i\right ] _+ 
      \end{aligned}
\end{equation}
in which $[\cdot]_+$ means "truncated to zero", that is, values less than zero are set to zero to ensure that $\alpha_n^j$ and $\alpha_p^i$ are non-negative numbers. $O_n$ is the best value of $s_j^n$, and $O_p$ is the best value of $s_i^p$. Under the cosine similarity metric, the target of $s_p$ is 1 and the target of $s_n$ is 0, so we can set $O_n=-m$, $O_p=1-m$, where $m$ is the margin.

Rescaling cosine similarity is a common practice in class-level label learning. In traditional methods, all similarity scores share the same scaling factor $\gamma$. Unlike traditional methods, circle loss multiplies each similarity score by an independent weighting factor before rescaling. This approach removes the restriction of equal rescaling and provides greater flexibility in the optimization process. In addition to better optimization effects, this reweighting strategy enables circle loss to hold a similar pair optimization perspective, making it compatible with class-level and pairwise label learning. Circle loss achieves both better within-class compactness and between-class discrepancy (on the training set), we believe that it indicates better optimization. 

\subsection{Adversarial training with circle loss}
To achieve adversarial robustness, we utilize a cross-entropy loss encouraging the model's predictions for the adversarial sample close to the true value and apply the circle loss to enhance the adversarial training effect, i.e., maximizing the within-class distance and minimizing the between-class distance. Thus the loss function of our algorithm is formulated as
\begin{equation}\label{eq:loss}
    \mathcal{L}=\mathcal{L}_{\mathrm{CE}}(\mathcal{F}(x^{\mathrm{adv}}),y)+\beta \cdot \mathcal{L}_{\mathrm{circle}}(\phi (x^{\mathrm{adv}}))
\end{equation}
where $\mathcal{L}_{\mathrm{CE}} (\cdot)$ is the cross-entropy loss, $ \mathcal{F}(x^{\mathrm{adv}})$ is the output vector of the learning model (with the softmax operator in the last layer), $\mathcal{L}_{\mathrm{{circle}}} (\cdot)$ is the circle loss, $\phi (x^{\mathrm{adv}})$ is the features of the samples extracted by the learning model, and $\beta>0$ is a scaling parameter that balances the two parts of the final loss.

The first term in Eq.~\eqref{eq:loss} guarantees the prediction accuracy of the model for adversarial examples by minimizing the difference between $\mathcal{F}(x^{\mathrm{adv}})$ and $y$, while the second regularization term enhances adversarial robustness, that is, it further pushes the decision boundary of the classifier away from the sample instances via maximizing the within-class distance and minimizing the between-class distance. The conceptual illustration is shown in~\autoref{fig:decisionboundary}.

\begin{algorithm}
\footnotesize 
\caption{Adversarial training with circle loss}
    \begin{algorithmic}[1]
        \STATE Initialize network $f(\theta)$;
        \REPEAT
        \STATE Read mini-batch $X$ from training set;
        \STATE Construct $X^{\mathrm{adv}}$ against $f(\theta)$ for each instance in $X$, adversarial attack methods choose from FGSM, PGD and C\&W;
        \STATE Train $f(\theta)$ with $X^{\mathrm{adv}}$ using Eq.~\eqref{eq:loss};
        \UNTIL {Training converged}
    \end{algorithmic}
\label{alg}
\end{algorithm}

The pseudocode of the adversarial training procedure is displayed in Algorithm~\ref{alg}.
In Line 4, we construct adversarial examples for adversarial training. 
As described in Eq.~\eqref{eq:advtrain}, the cost of adversarial training is dominated by solving the inner maximization problem. From this perspective, the success of learning a robust classifier depends on the quality of the inner maxima $x^{\mathrm{adv}}$. Unfortunately, direct optimization on Eq.~\eqref{eq:advtrain} is practically intractable due to the challenges in optimizing (nonconcave) inner maximization over all training data. In practice, we instead approximate the optimal adversary with a local maximum $x^{\mathrm{adv}}$. The most commonly used inner maximizer is the PGD algorithm, which can quickly generate aggressive adversarial examples. However, even if one method is chosen to generate high-quality local maxima $x^{\mathrm{adv}}$, the trained model may not be as robust to the adversarial examples generated by other adversarial methods. So, we attempt to combine different attack methods in adversarial training to increase the robustness. We consider three methods for generating adversarial examples, FGSM, PGD, and C\&W. These three methods also represent single-step attacks, iterative attacks, and optimization attacks, respectively. 
This increases the diversity of perturbations during training and can improve the robustness not only under a known type of attack but also under an unknown type of attack. Once we have the adversarial examples, we train the network by minimizing Eq.~\eqref{eq:loss} as Line 5 of Algorithm~\ref{alg}.

\section{Experiment}\label{S4}
In this section, we first introduce the SHM task, including datasets, and classifiers. 
Following this, we evaluate the performance of our method in scenarios involving white-box attacks, black-box attacks, and Gaussian noise.
Subsequently, we conduct ablation experiments to investigate the impact of circle loss on model robustness.

\subsection{SHM datasets and classifiers}\label{S4.1}

We selected two SHM datasets and established two distinct models based on their dataset characteristics: a simple ANN representing conventional machine learning methods and a complex network representing deep learning methods. Below are detailed descriptions of the datasets and classifiers used in this study.

\begin{figure*}[h]
    \centering
    \includegraphics[width=16cm]{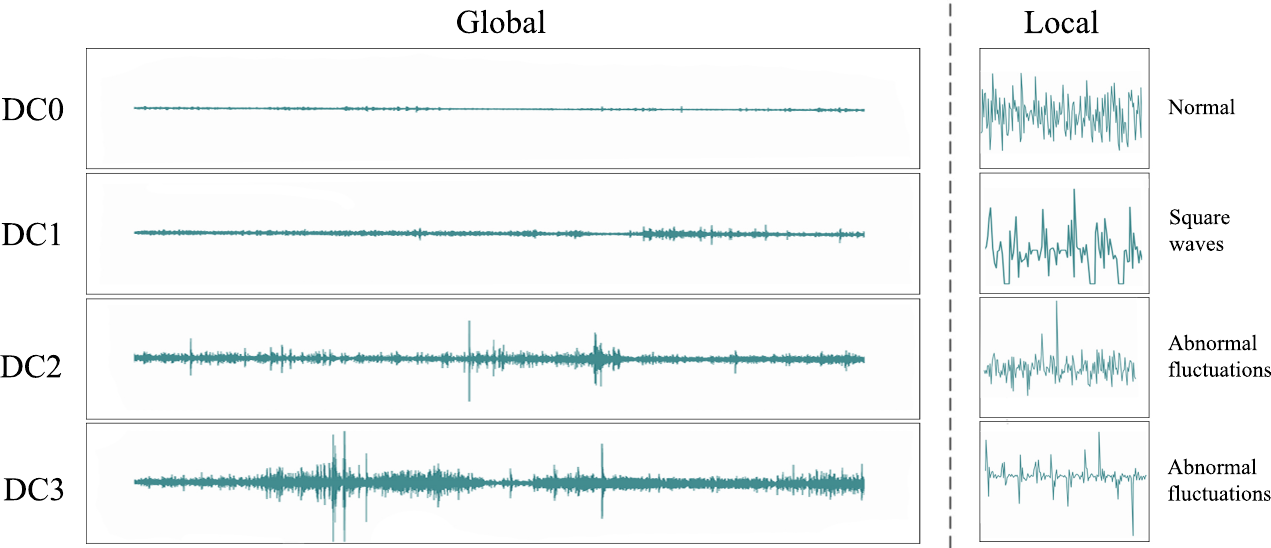}
\caption{Acceleration curves for four states and the typical samples for each state.}\label{fig:tcrfsample}
\end{figure*}

\begin{figure}[hbt]
    \centering
    \includegraphics[width=8.5cm]{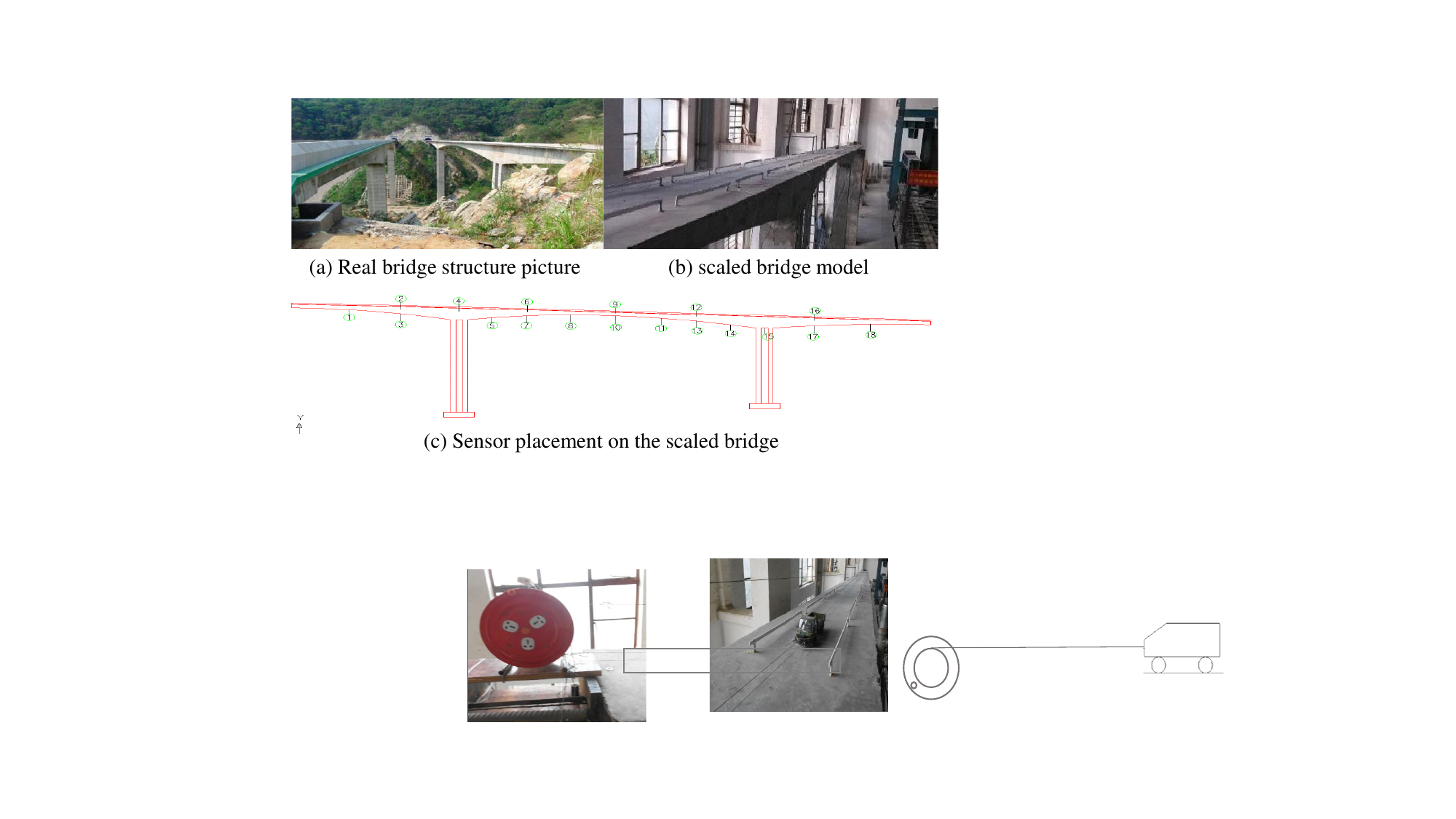}
    \caption{TCRF bridge scale model. \label{fig:hei}}
\end{figure}

\begin{figure}[h]
    \centering
    \includegraphics[width=6cm]{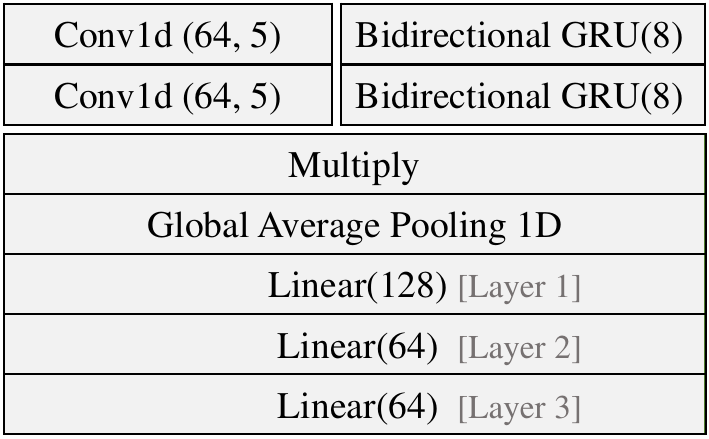}
    \caption{PCBG network structure.\label{fig:net}}
\end{figure}

\begin{figure*}[h]
    \centering
    \includegraphics[width=15.5cm]{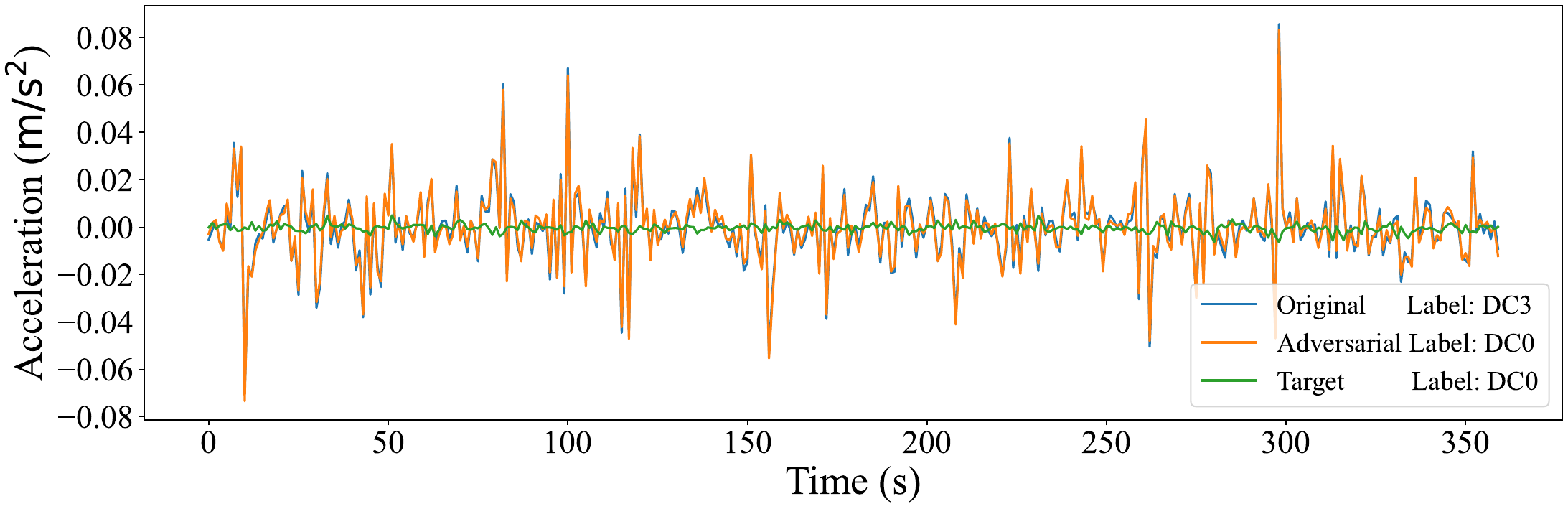}
    \caption{BIM attacks ($\epsilon=0.003$) on the TCRF bridge scale model dataset. The adversarial sample is semantically similar to the original samples of DC3 (damaged), but the classification results are the same as the target sample of DC0 (undamaged). }
    \label{fig:heiattack1}
\end{figure*}

\subsubsection{Three-span continuous rigid frame bridge (TCRF bridge) scale model}
We use a scale bridge model of the real TCRF bridge where the primary bridge structure, bridge pier, and bridge abutment are constructed following the same scaling ratio of 1:20~\cite{yang2020hierarchical}. 
The stiffness degradation of the bridge structure is simulated by applying the concentrated force in the span of the continuous rigid frame bridge to make floor cracks. We use these cracks to represent the structural damages. In our experiment settings, we have 4 kinds of structural damage states, which are shown in~\autoref{tab:casetcrf}. 

\begin{table}[h]
\footnotesize
  \centering
  \caption{Damage conditions (DC) of TCRF bridge scale model.}
    \begin{tabular}{ll}
    \toprule
    \makebox[0.11\textwidth][s]{Label} & \makebox[0.32\textwidth][s]{Descriptions} \\
    \midrule
    DC0   & No damage in the scale bridge  \\
    DC1   & One crack in the scale bridge \\
    DC2   & Two cracks in the scale bridge \\
    DC3   & Two larger cracks in the scale bridge \\
    \bottomrule
    \end{tabular}%
  \label{tab:casetcrf}%
\end{table}

\begin{figure*}[h]
    \centering
    \includegraphics[width=15cm]{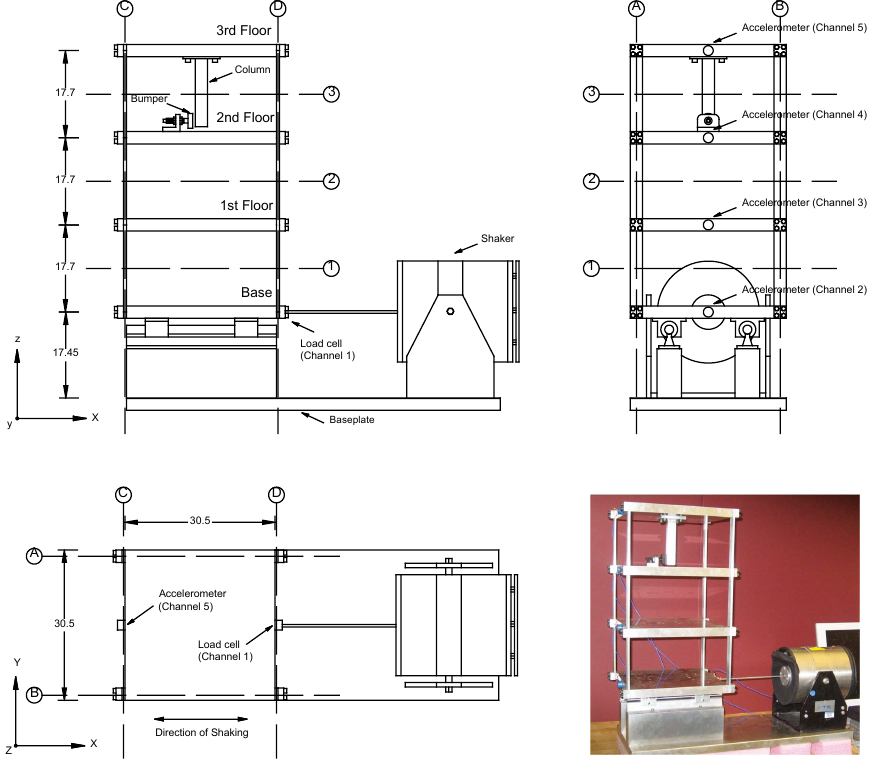}
    \caption{Basic dimensions of the three-storey building structure (all dimensions are in cm).\label{fig:threesketch}}
\end{figure*}

\begin{figure*}[h]
\centering
\subfigure[The samples of acceleration histories.]{
\begin{minipage}{0.45\linewidth}
\centering
\includegraphics[width=0.98\linewidth,height=6.5cm]{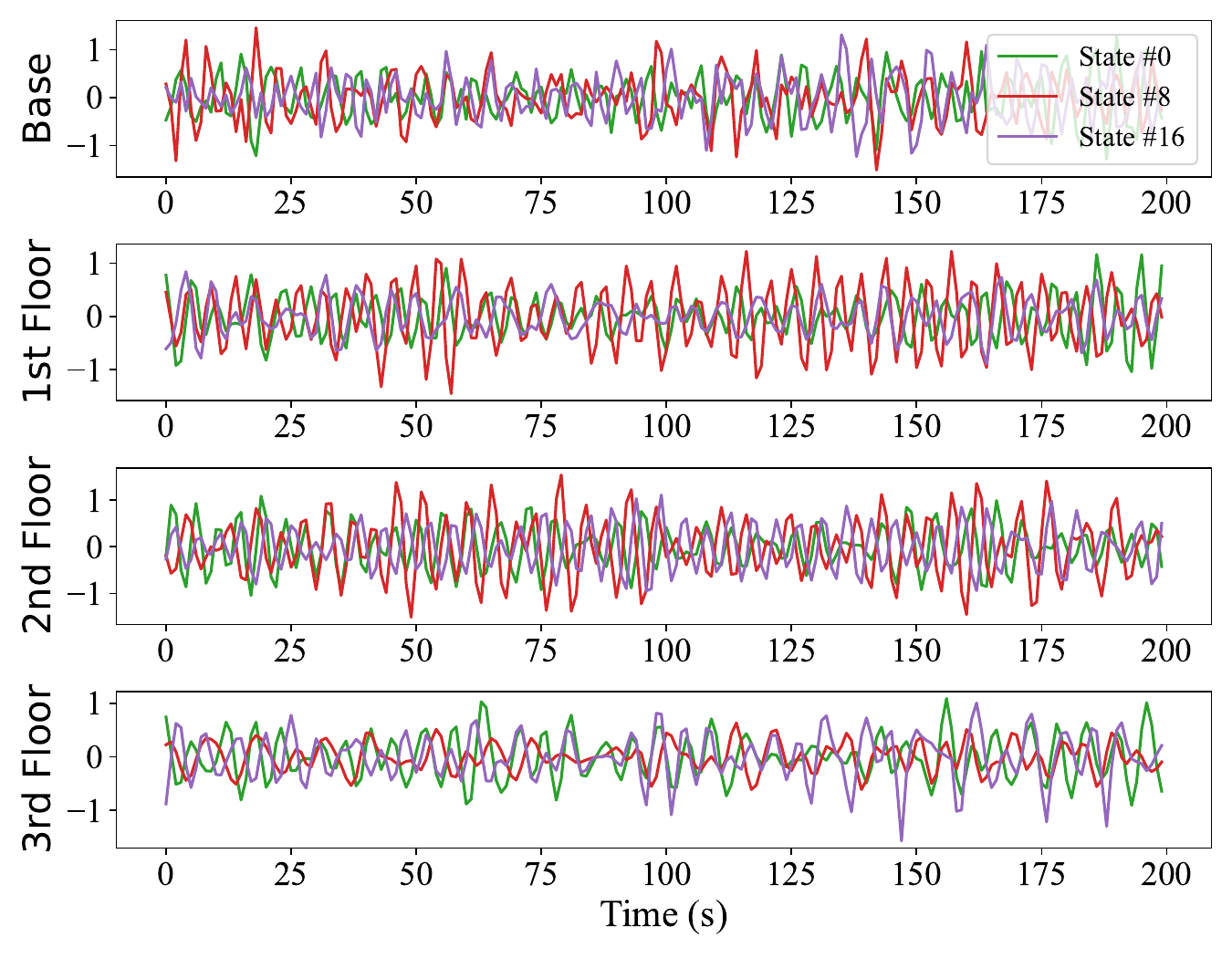}
\end{minipage}
}
\subfigure[The samples of FRFs.]{
\begin{minipage}{0.51\linewidth}
\centering
\includegraphics[width=0.98\linewidth,height=6.5cm]{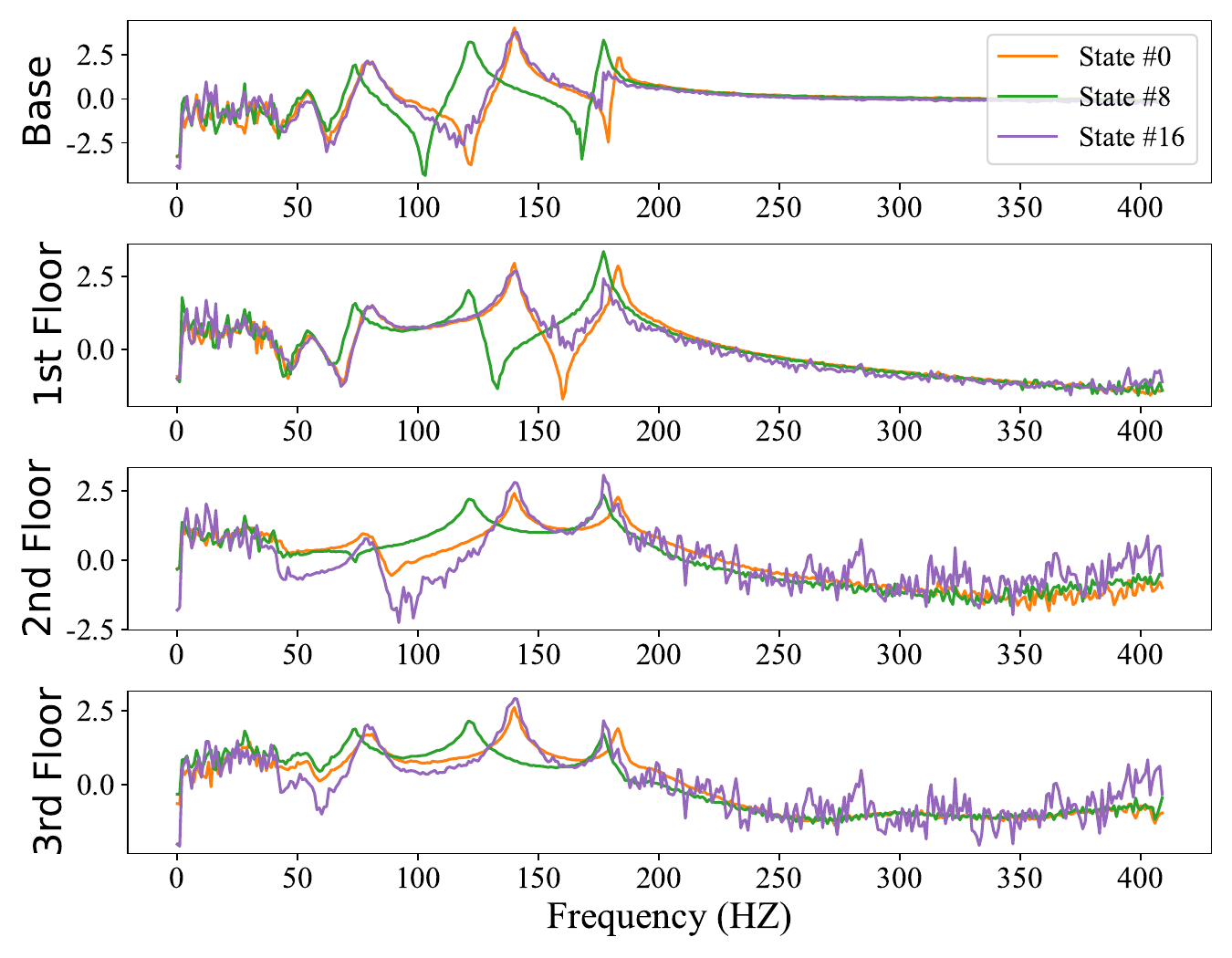}
\end{minipage}
}
\caption{Comparison of acceleration histories and FRFs for each floor of three samples belonging to different state conditions.}
\label{fig:three_samples}
\end{figure*}

\begin{figure*}[h]
    \centering
    \includegraphics[width=17.5cm,height=4.5cm]{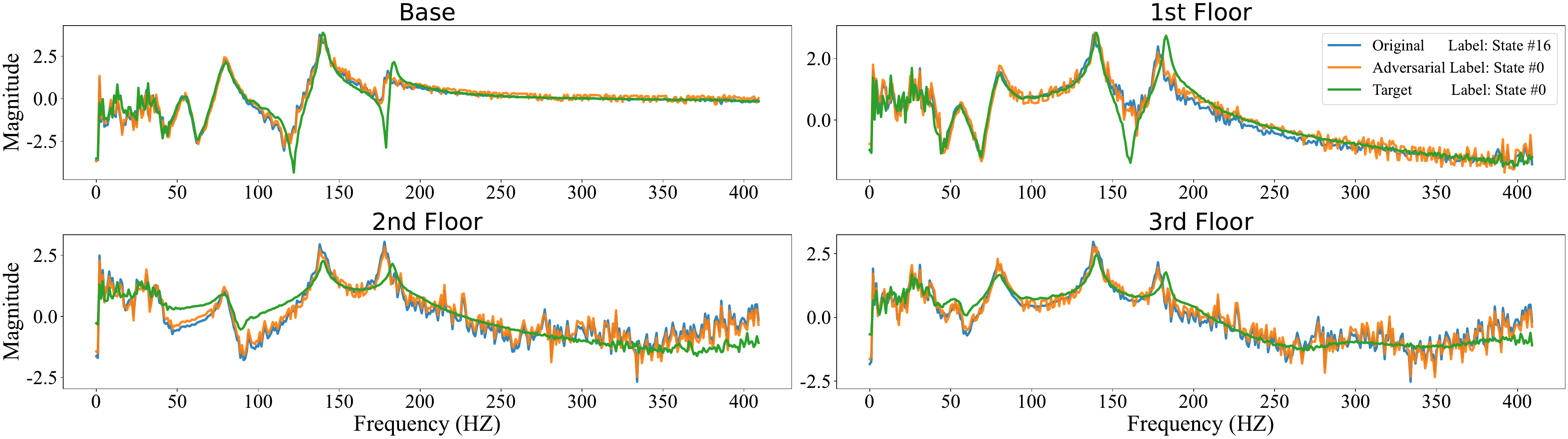}
    \caption{BIM attacks ($\epsilon=0.1$) on the LANL structure dataset. The adversarial sample is semantically similar to the original samples of State \#16 (damaged), but the classification results are the same as the target sample of State \#0 (undamaged). }
    \label{fig:threeattack1}
\end{figure*}

Dynamic characteristics of the structure itself such as the mode shape and natural frequency will change when a certain part of the structure is damaged. Among them, the vibration response acceleration information containing rich structure dynamic information can better reflect the overall damage to the structure, and the vibration response acceleration data is relatively easy to collect in actual engineering. 
So, to monitor the changes in the structural state degradation, 18 acceleration sensors have been installed on the scale model, including 12 vertical measuring points at the bottom of the beam and 6 on the web horizontal measuring points, as shown in \autoref{fig:hei}. 

We tow a 0.3kg scale car across the bridge deck in 15 seconds to simulate the dynamic load process under four structural damage states, then the acceleration can be recorded by the sensors at a sampling frequency of 8192 Hz. 
In \autoref{fig:tcrfsample}, it shows that when the car passes under different conditions, the data collected by the acceleration sensor changes significantly. For professionals, there is no need to process the data, and they can directly distinguish various state conditions from these acceleration data to obtain the health status of the structure. 

To obtain the best structural damage identification performance, our network structure borrows the parallel Convolutional Neural Network and Bidirectional Bidirectional Gated Recurrent Unit (PCBG) framework in~\cite{yang2021data}, which is currently a state-of-the-art network framework on this dataset.
The network structure and parameters are shown in~\autoref{fig:net}. 
The total number of data samples was 270000, each containing 320 data points (20 values for each of the 18 sensors), all of which were fed into this network for damage recognition classification.
The network was trained by dividing the dataset into a training set and a validation set in the ratio of 7:3, and the network was able to achieve an accuracy of 99.51\%  on the entire dataset.

The neural network exhibits robust accuracy on the dataset, maintaining a 94.03\% accuracy even when the dataset is augmented with Gaussian noise of mean amplitude 0.003. However, when adversarial noise generated by the BIM adversarial attack with a perturbation magnitude of 0.003 is introduced into the dataset, the model's accuracy drops drastically to 26.75\% (as depicted in a typical case in~\autoref{fig:heiattack1}). This demonstrates a significantly more severe impact of adversarial noise compared to Gaussian noise at an equivalent level, highlighting the heightened threat posed by adversarial perturbations.

\begin{table}[h]
\footnotesize 
 \renewcommand{\arraystretch}{0.9}
  \centering
  \caption{Structural damage states of the three-storey structure.}
  \resizebox{\linewidth}{!}{
    \begin{tabular}{ll}
    \toprule Label  & Description\\
    \midrule
    State \#0  & Baseline condition\\
    State \#1  & Mass=1.2kg at the base\\
    State \#2  & Mass=1.2kg on the 1st floor\\
    State \#3  & 87.5\% stiffness reduction in column 1BD\\
    State \#4  & 87.5\% stiffness reduction in column 1AD and 1BD\\
    State \#5  & 87.5\% stiffness reduction in column 2BD\\
    State \#6  & 87.5\% stiffness reduction in column 2AD and 2BD\\
    State \#7  & 87.5\% stiffness reduction in column 3BD\\
    State \#8  & 87.5\% stiffness reduction in column 3AD and 3BD\\
    State \#9  & Gap=0.20mm\\
    State \#10  & Gap=0.15mm\\
    State \#11  & Gap=0.13mm\\
    State \#12  & Gap=0.10mm\\
    State \#13  & Gap=0.05mm\\
    State \#14  & Gap=0.20mm and mass=1.2kg at the base\\
    State \#15  & Gap=0.20mm and mass=1.2kg on the 1st floor\\
    State \#16  & Gap=0.10mm and mass=1.2kg on the 1st floor\\
    \bottomrule
    \end{tabular}%
    }
  \label{tab:casethree}%
\end{table}

\subsubsection{Los Alamos National Laboratory (LANL) three-storey structure} 

The basic dimensions of the three-storey structure~\cite{figueiredo2009structural} are shown in~\autoref{fig:threesketch}.
There are 17 structural state conditions, information that describes the different states as shown in~\autoref{tab:casethree}. For example, the state condition labeled "State \#3" is described as "87.5\% stiffness reduction in column 1BD", which means there was 87.5\% stiffness reduction in the column located between the base and first floor at the intersection of plane B and D as defined in~\autoref{fig:threesketch}. The "gap" mentioned in the descriptions of States \#9–\#16 means the distance between the bumper and the suspended column, which is variable and used to introduce different levels of nonlinearity for a given level of excitation.
The structure is excited with a band-limited (20–150 Hz) Gaussian forcing signal by an electrodynamic shaker attached to the base, and one sensor per layer is used to collect the force and acceleration histories under various structural state conditions. Each data sample for SHM contains excitation signal as well as the acceleration response information for the base, the 1st floor, the 2nd floor, and the 3rd floor of the structure.

In~\autoref{fig:three_samples}, State \#0 is the undamaged condition, and State \#16 is the damaged condition, but their force and acceleration histories plots are displayed too similarly to be distinguished by the eye. Considering that State \#0 and State \#16 belong to two completely different state conditions, the difference they show should be large enough to be noticed. So we use the ratio of acceleration to excitation force to extract the FRFs of the data.
FRFs are frequently used as damage and condition-sensitive features in SHM as they encode physical dynamic properties and can be efficiently computed. Their peaks and valleys have been revealed to be important features for damage characterization. 
To a trained engineer, FRFs of a structure are approximately tractable visually, and structural damage information can be easily derived from the offset of their peaks and valleys~\cite{chesne2013damage}. 
As shown in~\autoref{fig:three_samples}b, the normalized FRFs of different state conditions show shifts of the resonance frequencies as well as distortions of the FRF shape caused by damage.
The FRFs were estimated for each layer of each sample using the non-overlapping Welch method, Hanning windows, and five-fold averaging. 
After extracting FRF features, our classification task is a lot lighter and can be done with a simple ANN. We set up a three-layer ANN containing an input layer, hidden layer, and output layer, where the hidden layer contains 17 nodes. 
The dataset contains 1700 samples with 1640 data points per sample. The network was trained by dividing the dataset into a training set and a validation set in the ratio of 7:3, and the network was able to achieve an accuracy of 99.60\% on the entire dataset.

\begin{table*}[ht]
\footnotesize 
\caption{ Natural accuracy and parameter settings of defense models.}
\label{tab:parameters}
\centering
\begin{tabular}{llcl}
\toprule
 & \makebox[0.1\textwidth]{} & \makebox[0.16\textwidth]{Natural Accuracy (\%)} & \makebox[0.53\textwidth][s]{Parameters} \\ 
 \midrule
 \multirow{1}{*}{TCRF Bridge}
 &RS & 97.94 & Failure probability = 0.1 and noise level hyperparameter = 0.003 \\
 & Distillation & 98.93 & Temperature = 2048 and balancing factor = 0.4 \\
 & Fast-AT & 98.86 & Radius = 0.003 and step size = 0.0045 \\
 & PGD-AT & 94.81 & PGD with 20 steps and step size = 0.0003 \\
 \midrule
\multirow{1}{*}{LANL Structure} 
 & RS & 99.51 & Failure probability = 0.1 and noise level hyperparameter = 0.1 \\
 & Distillation & 98.29 & Temperature = 2 and balancing factor = 0.85 \\
 & Fast-AT & 97.59 & Radius = 0.5 and step size = 0.75 \\
 & PGD-AT & 98.47 & PGD with 20 steps and step size = 0.005 \\
\bottomrule
\end{tabular}
\end{table*}

The neural network exhibits robust accuracy on the dataset, maintaining a 98.94\% accuracy even when the dataset is augmented with Gaussian noise of mean amplitude 0.1. However, when adversarial noise generated by the BIM adversarial attack with a perturbation magnitude of 0.1 is introduced into the dataset, the model's accuracy drops drastically to 6.53\% (as depicted in a typical case in~\autoref{fig:threeattack1}). This demonstrates a significantly more severe impact of adversarial noise compared to Gaussian noise at an equivalent level, highlighting the heightened threat posed by adversarial perturbations.

\subsection{Performance under white-box attacks}
In our approach, $L_{\mathrm{circle}}$ operates on the penultimate layer features. The following transformation of the penultimate layer only consists of a linear layer and a softmax layer, which ensures that small fluctuations in the embedding will only result in a monotonous adjustment to the output controlled by some tractable Lipschitz constant~\cite{mao2019metric,oberman2018lipschitz}. The penultimate layer tends to preserve more information than the logit layer. 
For the TCRF bridge dataset, we configured the proportion of adversarial examples in training as PGD: FGSM: C\&W = 3: 1: 1, employing Equation ~\ref{eq:loss} with a scaling parameter $\beta$ of 0.01. The resultant model achieved an accuracy of 96.98\% on the clean dataset.
Regarding the LANL structure dataset, the training comprised a proportion of adversarial examples set to PGD: FGSM: C\&W = 1: 3: 1, utilizing Equation~\ref{eq:loss} with a scaling parameter $\beta$ of 0.1. The model attained an accuracy of 95.88\% on the clean dataset.

We confirm the efficacy of our defense method by conducting identical experimental procedures on the aforementioned TCRF bridge and LANL structure datasets. Initially, we compare the defense models trained by our approach against the original models mentioned in~\autoref{S4.1} under BIM and FGSM attacks, presenting the outcomes in~\autoref{tab:CompareWithStandard}.

\begin{table}[h]
\footnotesize 
\renewcommand{\arraystretch}{1.1}
\setlength{\tabcolsep}{3pt}
\caption{Comparison of the accuracy of the model trained using our proposed defense method and the original model under FGSM and BIM attack}
\label{tab:CompareWithStandard}
\centering
\begin{tabular}{llcccc}
\toprule
 \multirow{2}{*}{} & \multirow{2}{*}{}
                        & \multicolumn{2}{c}{Standard } & \multicolumn{2}{c}{Our}\\ 
                        \cmidrule(r){3-4}  \cmidrule(r){5-6}
                       &&\makebox[0.06\textwidth]{FGSM} & \makebox[0.06\textwidth]{BIM} &\makebox[0.06\textwidth]{FGSM} & \makebox[0.07\textwidth]{BIM} \\ \midrule
\multirow{3}{*}{TCRF Bridge}
                      &$\epsilon=0.001$ &65.11 &57.42 &  94.38 &  90.56 \\
                      &$\epsilon=0.003$ &38.14 &26.75 & 86.98 &   81.08 \\
                      &$\epsilon=0.005$ &27.46 &14.76 &  65.63   & 57.15  \\ \midrule
\multirow{3}{*}{LANL Structure} 
                      &$\epsilon=0.05$ &61.53 &58.59 &  89.35 &   85.88  \\
                      &$\epsilon=0.1$  &13.76 &6.53 &  77.18    &   76.47  \\
                      &$\epsilon=0.15$  &0.29  &0.00 & 57.41    &     54.43   \\ \bottomrule
\end{tabular}
\end{table}

As the perturbation magnitude $\epsilon$ increases, the attack becomes more potent, causing a quicker decline in model accuracy.
Our defense model showcases superior resilience, maintaining higher accuracy levels against adversarial attacks compared to standard models. Particularly noteworthy is the ability of our method to sustain over half of its accuracy in the LANL structure dataset at $\epsilon=0.15$, while the original model's accuracy drops to zero. Furthermore, our defense approach exhibits notable resilience against smaller perturbations, preserving significantly higher accuracy levels than standard models. These findings underscore the effectiveness of our method in fortifying adversarial robustness.

To rigorously assess the efficacy of our defense approach, we subsequently compare it against four well-established and effective defense methodologies introduced in \autoref{S2.3}, i.e., Randomized Smoothing (RS), Distillation, Fast adversarial training (Fast-AT) and
PGD-based adversarial training (PGD-AT).

For all methods, we use the same network architectures that are specified in ~\autoref{S4.1} and turn the parameters to ensure they show the best performance, the natural accuracy and parameter settings of these defense models are shown in~\autoref{tab:parameters}.  

In addition to BIM and FGSM, which are gradient-based attack methods, we incorporate C\&W and AdvGAN attacks. Unlike BIM and FGSM, C\&W and AdvGAN are respectively based on optimization and generator techniques.
All attacks have full access to model parameters and are constrained by the same perturbation limit $\epsilon$.
The added adversarial noise maintains a signal-to-noise ratio between 25-50 dB concerning the original sample, aligning with typical noise levels present in SHM systems~\cite{PrabakaranComparison, campeiro2018impedance, flynn2010bayesian}.
In the C\&W attack in the TCRF bridge scenario, the number of iterations is 1000, the misclassification confidence factor $k=0$, and the objective function's weight ~$c=0.0001$. In the C\&W attack in the LANL structure scenario, the number of iterations is 1000, the misclassification confidence factor $k=0$, and the objective function's weight ~$c=2$. All C\&W attacks use the $L_2$ attack version.

\begin{table}[h]
\footnotesize 
\renewcommand{\arraystretch}{0.85}
\setlength{\tabcolsep}{4.6pt}
\caption{White-box robustness (accuracy (\%) on white-box test attacks) on TCRF bridge and LANL structure scenarios.}
\label{tab:white}
\centering
\begin{tabular}{lcccccc}
\toprule
&Standard & RS&Distillation & Fast-AT & PGD-AT & Our \\
\midrule
 \multicolumn{7}{c}{TCRF Bridge $(\epsilon=0.003)$}\\
\midrule
   FGSM & 38.14 & 43.19 & 49.36 & 84.13 & 86.88 & \textbf{86.98} \\
  BIM & 26.75 & 29.23 & 33.94 & 79.24 & 80.30 & \textbf{81.08} \\
  C\&W & 1.19 & 17.15 & 1.02 & 9.76 & 25.19 & \textbf{25.22} \\
  AdvGAN & 71.31 & 73.76 & 73.83 & 65.64 & 91.49 & \textbf{93.85} \\ 
 \midrule
 \multicolumn{7}{c}{LANL Structure $(\epsilon=0.1)$}\\
  \midrule
   FGSM & 13.76 & 36.25 & 38.12 & 66.82 & 70.76 & \textbf{77.18} \\
  BIM & 6.53 & 29.31 & 33.65 & 63.00& 69.47 & \textbf{76.47} \\
  C\&W              & 0.00 & 0.00 & 38.94 &16.88 & 25.24 & \textbf{55.47} \\
  AdvGAN & 41.12 & 41.31 & 55.06 & 80.82 & 83.76 & \textbf{84.29} \\
\bottomrule
\end{tabular}
\end{table}

The white-box robustness of all defense models is reported in ~\autoref{tab:white}. 
our defense mechanism shows the highest robustness against FGSM, BIM, C\&W, and AdvGAN attacks on both the TCRF bridge and LANL structure SHM datasets. Particularly notable is the significant enhancement in adversarial robustness against these attacks on the LANL structure scenario compared to the second-best approach. Our method achieves approximately a $7\%$ improvement in robustness against FGSM and BIM attacks, while demonstrating a substantial improvement of approximately $30\%$ against C\&W attacks, showcasing its remarkable effectiveness in mitigating diverse adversarial attacks.

\subsection{Performance under black-box attacks} 
\begin{table}[h]
\footnotesize 
\renewcommand{\arraystretch}{0.85}
\setlength{\tabcolsep}{4.6pt}
\caption{Black-box robustness (accuracy (\%) on white-box test attacks) on TCRF bridge and LANL structure scenarios.}
\label{tab:black}
\centering
\begin{tabular}{lcccccc}
\toprule
&Standard & RS&Distillation & Fast-AT & PGD-AT & Our \\
\midrule
 \multicolumn{7}{c}{TCRF Bridge $(\epsilon=0.003)$}\\
\midrule
 FGSM & 76.52 & 81.28 & 77.32 & 86.64 & 94.63 & \textbf{95.55} \\
 BIM$^{20}$ & 75.46 & 81.01 & 78.34 & 82.60 & 92.52 & \textbf{93.98} \\
 C\&W & 93.74 & 97.14 & 92.78 & \textbf{98.16} & 92.64 & 95.51 \\
 AdvGAN & 67.78 & 69.94 & 71.04 & 84.02 & 92.17 & \textbf{92.97} \\
 \midrule
 \multicolumn{7}{c}{LANL Structure $(\epsilon=0.1)$}\\
 \midrule
 FGSM & 50.82 & 51.89 & 54.59 & 65.41 & 77.12 & \textbf{82.00} \\
 BIM$^{20}$ & 44.82 & 44.56 & 52.88 & 64.00 & 77.35 & \textbf{81.88} \\
 C\&W & 75.76 & 79.86 & 81.94 & 94.47 & 92.53 & \textbf{92.65} \\
 AdvGAN & 66.29 & 66.72 & 64.18 & 81.29 & 85.41 & \textbf{88.00} \\
\bottomrule
\end{tabular}
\end{table}

Black-box attacks are executed using a substitute model approach, harnessing the transferability of adversarial examples to target the model. The process involves generating synthetic datasets, training substitute models, and conducting transfer attacks.
In the TCRF bridge SHM scenario, the chosen substitute model is the HCG model~\cite{yang2020hierarchical}, an innovative hierarchical framework that amalgamates Convolutional Neural Networks and Gated Recurrent Units to effectively capture spatial and temporal relations.
In the LANL three-storey structure scenario, the substitute model is an ANN with a hidden layer comprising 32 nodes.

In the black-box scenario, the adversaries are still FGSM, BIM$^{20}$, C\&W, and AdvGAN. The black-box robustness of all defense strategies is reported in~\autoref{tab:black}. Compared with the white-box results, all defense methods achieve much better robustness against black-box attacks, even close to the natural accuracy. Once again, our defense model demonstrates superior robustness compared to other baseline approaches.

\subsubsection{Transferability test}
We investigate the transferability of attacks on the TCRF bridge dataset and LANL structure dataset between a standard training model, distillation model, Fast-AT model, PGD-AT model, and our model.  
In~\autoref{tab:trans1} to~\ref{tab:trans4}, we present the accuracy of target models (columns) when subjected to attacks from adversarial samples generated from source models (rows). The bottom rows indicate the average accuracy of the target model against adversarial samples transferred from other source models.

In comparison to other models, our model showcases remarkable resilience against adversarial samples transferred from other models. Particularly on the LANL dataset, our model boasts an average accuracy approximately 7\% higher than the second-best model when confronted with adversarial samples generated by other models

\begin{table}[h]
\footnotesize
\setlength{\tabcolsep}{5pt}
\caption{Transferability test on TCRF bridge dataset: FGSM adversaries are generated with $\epsilon=0.003$ using the source network and then evaluated on the target model. }
\centering
\label{tab:trans1}
\begin{tabular}{c|ccccc}
\toprule
\diagbox{Source}{Target} &Standard &Distillation & Fast-AT & PGD-AT & Our\\
\midrule
Standard&38.12&47.82&91.72&92&93.06\\
Distillation&52.28&49.36&92.61&93.03&94.01\\
Fast-AT&80.31&79.14&84.13&87.87&89.52\\
PGD-AT&81.96&82.76&87.53&86.88&90.03\\
Our&78.55&77.95&87.45&88.35&86.98\\ \midrule
Mean&66.24&67.41&88.69&89.63&\textbf{90.72}\\
\bottomrule
\end{tabular}
\end{table}

\begin{table}[h]
\footnotesize
\setlength{\tabcolsep}{5pt}
\caption{Transferability test on TCRF bridge dataset: BIM adversaries are generated with $\epsilon=0.003$ using the source network and then evaluated on the target model. }
\label{tab:trans2}
\centering
\begin{tabular}{c|ccccc}
\toprule
\diagbox{Source}{Target} &Standard &Distillation & Fast-AT & PGD-AT & Our\\
\midrule
Standard&26.74&37.63&92.21&92.42&94.15\\
Distillation&39.89&33.94&93.39&93.63&94.88\\
Fast-AT&79.94&79.23&79.24&86.42&87.65\\
PGD-AT&82.46&84.47&86.43&83.29&88.49\\
Our&75.00&74.98&86.32&87.28&81.08\\ \midrule
Mean&60.81&62.05&87.52&88.61&\textbf{89.25}
\\ \bottomrule
\end{tabular}
\end{table}

\begin{table}[h]
\footnotesize
\setlength{\tabcolsep}{5pt}
\caption{Transferability test on LANL structure dataset: FGSM adversaries are generated with $\epsilon=0.003$ using the source network and then evaluated on the target model.}
\label{tab:trans3}
\centering
\begin{tabular}{c|ccccc}
\toprule
\diagbox{Source}{Target} &Standard &Distillation & Fast-AT & PGD-AT & Our\\
\midrule
Standard&13.76&45.12&72.41&76.12&82.71\\
Distillation&29.76&38.12&72.71&75.94&81.59\\
Fast-AT&63.24&68.00&66.82&71.65&80.59\\
PGD-AT&62.94&69.24&68.24&70.76&80.82\\
Our&56.94&61.71&69.94&73.24&77.18\\ \midrule
Mean&45.33&56.44&70.02&73.54&\textbf{80.58}\\
 \bottomrule
\end{tabular}
\end{table}

\begin{table}[h]
\footnotesize
\setlength{\tabcolsep}{5pt}
\caption{Transferability test on LANL structure dataset: BIM adversaries are generated with $\epsilon=0.003$ using the source network and then evaluated on the target model.}
\label{tab:trans4}
\centering
\begin{tabular}{c|ccccc}
\toprule
\diagbox{Source}{Target} &Standard &Distillation & Fast-AT & PGD-AT & Our\\
\midrule
Standard&6.53&42.35&71.29&75.71&82.00\\
Distillation&24.65&33.65&72.65&76.59&81.65\\
Fast-AT&58.94&64.47&63.00&70.53&79.29\\
PGD-AT&58.47&65.71&65.88&69.47&79.71\\
Our&56.29&59.76&69.82&73.06&76.47\\ \midrule
Mean&40.98&53.19&68.53&73.07&\textbf{79.82}\\
 \bottomrule
\end{tabular}
\end{table}

The research by~\cite{kurakin2016adversarial,xie2019improving} suggests a trend where iterative attacks often overly adapt to specific network parameters, leading to high success rates in white-box scenarios but diminished performance in black-box scenarios. On the other hand, single-step attacks tend to have lower success rates in white-box settings but yield slightly better transferability of adversarial examples.
However, regarding transferability, our experimental findings differed from this trend as we observed that BIM adversarial examples demonstrated stronger transferability compared to FGSM adversarial examples. We believe that the evaluation of transferability might be notably influenced by the dataset used.

\subsection{Performance under Gaussian Noise}
A robust defense method should not only counter meticulously designed adversarial noise but also demonstrate resilience against the inherent and inevitable random white noise present in SHM systems. Although white noise of the same magnitude does not possess the same attacking power as adversarial noise, it can still disrupt machine learning models.
Numerous articles investigate the impact of white noise on SHM systems. For example, Campeiro et al.~\cite{campeiro2018impedance} present an experimental analysis of white noise effects on structural damage detection in impedance-based SHM systems, they indicate that even a low noise causes significant variations in the impedance signatures. Balasubramanian et al.~\cite{PrabakaranComparison} prove the inherent noise present in the sensor response poses a substantial hurdle for SHM systems based on neural networks to estimate the external impact correctly. 

To emulate environmental loads such as aerodynamic pressure, wind disturbances, and electrical noise, we intentionally injected Gaussian noise following a $\mathcal{N}(0, \sigma^2)$ distribution into the data. This was done to explore the effectiveness of our method against Gaussian noise. The results are presented in~\autoref{tab:gaussian}.

\begin{table}[h]
\footnotesize 
\setlength{\tabcolsep}{9pt}
\caption{Comparison of the accuracy of the model trained using our proposed defense method and the original model under Gaussian noise.}
\label{tab:gaussian}
\centering
\begin{tabular}{llcc}
\toprule
&Noise Level&\makebox[0.08\textwidth]{Standard}  & \makebox[0.08\textwidth]{Our} \\  \midrule
\multirow{3}{*}{TCRF Bridge}
                    &$\sigma=0.001$ &99.42&97.07\\
                    &$\sigma=0.003$ &94.03&97.19\\
                    &$\sigma=0.005$ &80.54&97.28\\
                    &$\sigma=0.007$ &69.53&86.00\\
                    \midrule
\multirow{3}{*}{LANL Structure} 
                    &$\sigma=0.4$ &96.59&94.29\\
                    &$\sigma=0.6$ &91.76&94.41\\
                    &$\sigma=0.8$ &88.47&92.88\\
                    &$\sigma=1.0$ &84.76&92.82\\
                    \bottomrule
\end{tabular}
\end{table}
It is evident from the table that the presence of Gaussian noise impacts the accuracy of SHM ML models. However, it's noteworthy that our method's trained models showcase remarkable resilience to Gaussian noise. The accuracy of our models on samples affected by Gaussian noise remains consistently similar to the accuracy achieved on clean, uncontaminated samples. This ability to maintain comparable accuracy levels in the presence of noise signifies the robustness of our models and renders them well-suited for deployment in SHM systems where noise interference is inevitable.
Furthermore, an intriguing observation is that in the TCRF bridge dataset, our model's accuracy on samples contaminated with Gaussian noise (with $\sigma$ values of 0.003 and 0.005) is marginally higher compared to the accuracy on clean samples. 
The addition of these Gaussian noises can even be likened to a beneficial form of smoothing preprocessing for our model's input data in certain scenarios.

\subsection{Ablation studies}
\subsubsection{$\mathcal{L}_{\mathrm{circle}}$ at different layers}
As the TCRF bridge scenario involves the application of a deep neural network model, we delved into the impact of employing circle loss at different depths within this context. Specifically, after each layer identified in ~\autoref{fig:net}, we applied circle loss to regulate the distances between features. Subsequently, we evaluated the defensive efficacy of these models post-training and reported their performance against FGSM and BIM attacks in~\autoref{tab:layer}.

\begin{table}[hpbt]
\footnotesize
\setlength{\tabcolsep}{12.pt}
\caption{Ablation analysis with $\mathcal{L}_{\mathrm{circle}}$ applied at different layers of PCBG network (~\autoref{fig:net}) for TCRF bridge dataset.}
\centering
\label{tab:layer}
\begin{tabular}{cccc}
\toprule
\multicolumn{1}{c}{Layer} & \multicolumn{1}{c}{No Attack} & \begin{tabular}[c]{@{}c@{}}FGSM\\ $(\epsilon=0.003)$\end{tabular} & \begin{tabular}[c]{@{}c@{}}BIM\\ $(\epsilon=0.003)$\end{tabular} \\ \midrule
None        &97.32&84.81&78.29     \\
Layer 1     &96.85&86.01&80.49     \\
Layer 2     &96.98&86.98&81.08     \\
Layer 3     &96.87&85.35&77.78     \\
Layer 1+2   &96.90&85.56&80.01     \\
Layer 1+3   &96.19&80.55&75.75     \\
Layer 2+3   &97.07&86.00&79.77     \\
Layer 1+2+3 &96.96&85.98&81.29     \\ \bottomrule
\end{tabular}
\end{table}

Training without the integration of circle loss can be likened to softmax adversarial training using mixed adversarial samples. The model trained in this manner showcases a marginal improvement in accuracy compared to models trained exclusively on PGD adversarial samples.

As can be seen, opting to implement circle loss after the penultimate layer proves to be a favorable choice. The model trained in this manner exhibits relatively high original accuracy and accuracy against FGSM and BIM attacks.

\subsubsection{Circle regularization} 
\begin{figure}[hpbt]
\centering
\subfigure[TCRF bridge.]{
\begin{minipage}{4.20cm}
\centering
\includegraphics[scale=0.16]{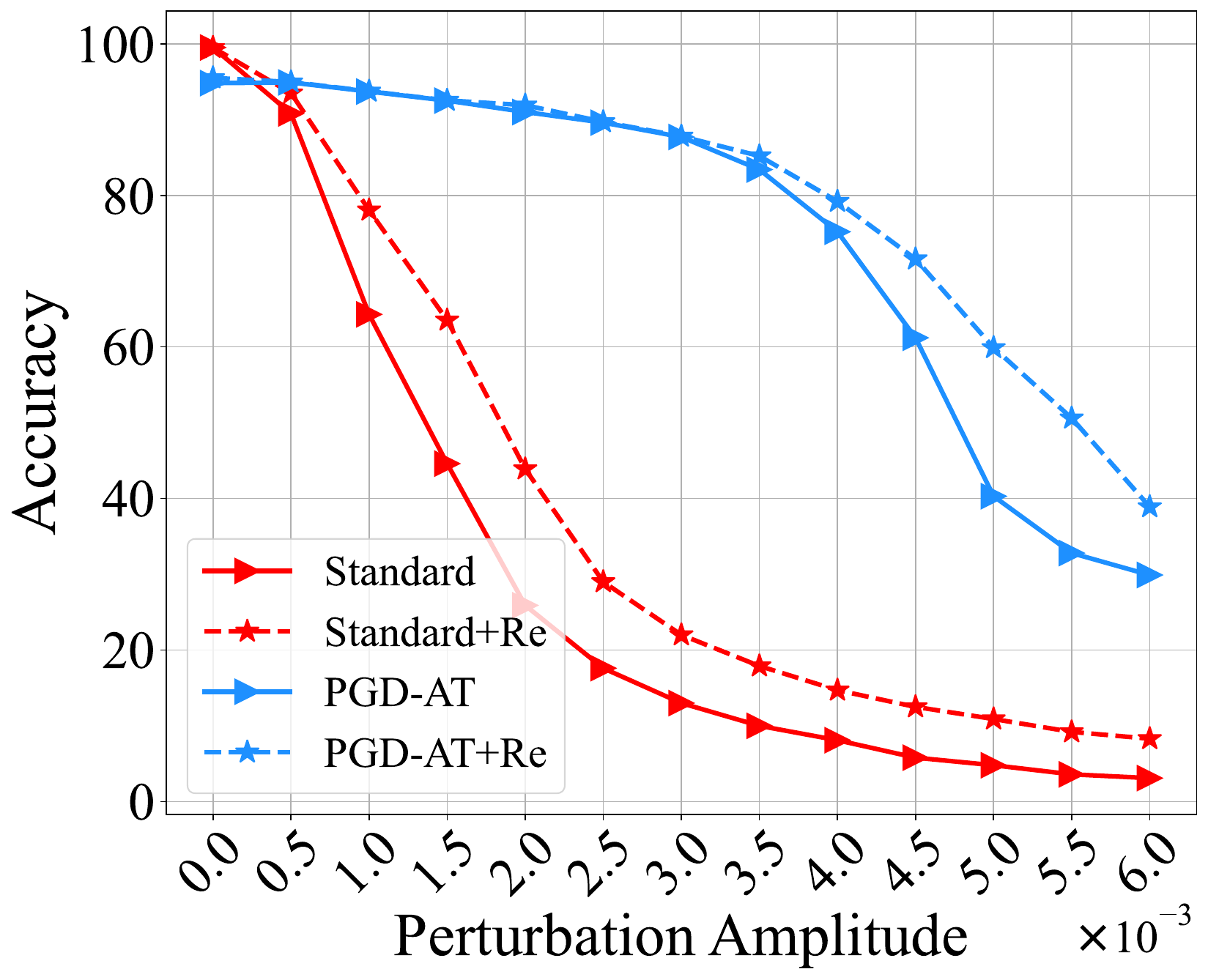}
\end{minipage}
}
\subfigure[LANL structure.]{
\begin{minipage}{4.20cm}
\centering
\includegraphics[scale=0.16]{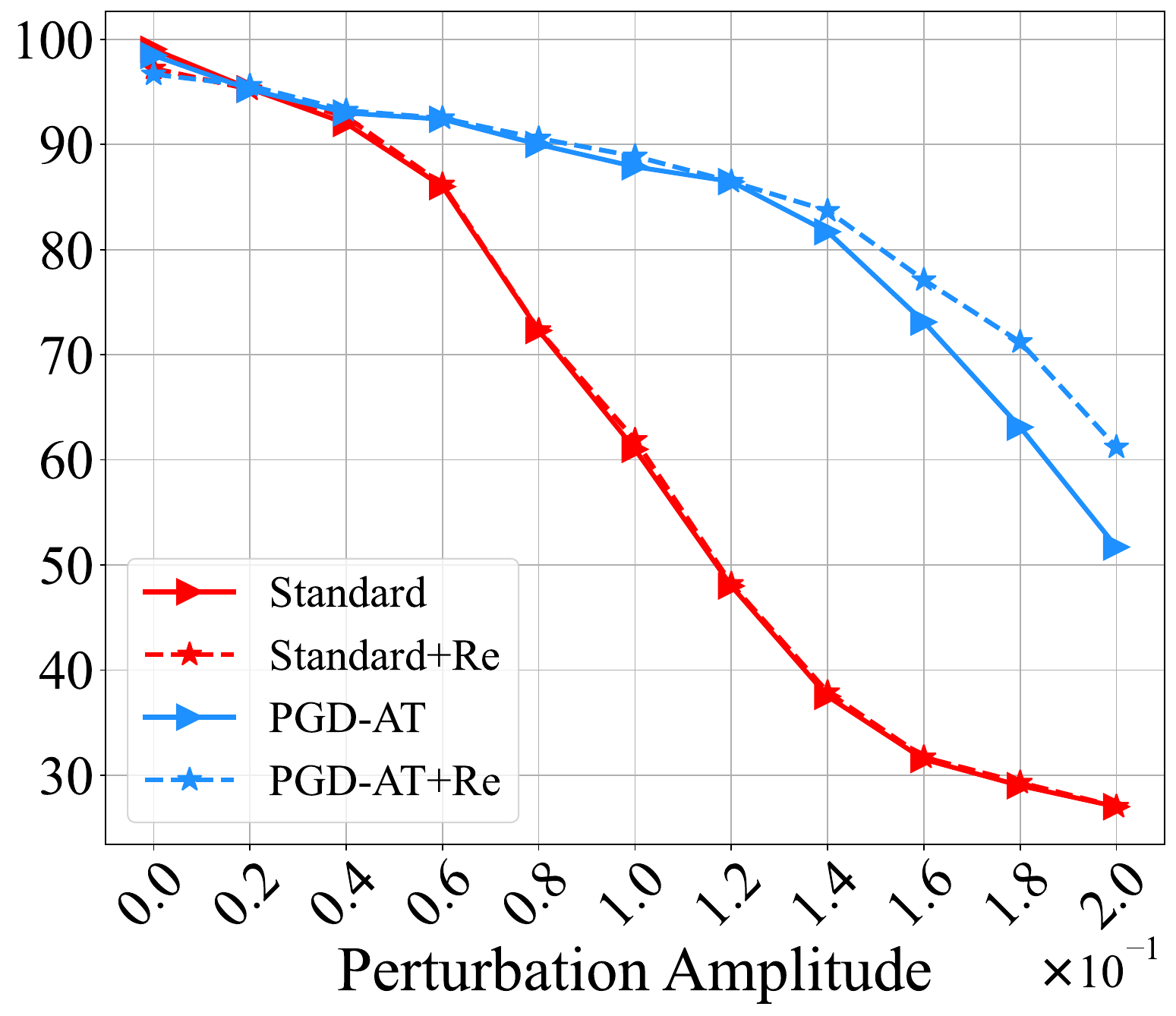}
\end{minipage}
}
\caption{Impact of regularization term. The BIM attacks on two datasets, ‘Standard’ and ‘PGD-AT’ means traditional training and PGD adversarial training methods, and ‘Re’ means adding circle loss regularization term to the training loss.}
\label{fig:regularization}
\end{figure}

To further investigate the enhancement of model robustness due to circle loss, we compare the standard training with and without circle loss regularization terms. Additionally, we explore the comparison between PGD adversarial training with and without circle loss regularization terms.
As shown in ~\autoref{fig:regularization}, models trained using the circle regularization term can be resistant to BIM attacks with a larger perturbation budget.

The experimental results indicate that circle loss can also be regarded as a regularization term. Thus, it can be incorporated into the standard training process to reduce the vulnerability of the model to adversarial examples and can also be incorporated into most of the existing defense methods for better robustness.

\section{Conclusion}\label{S5}

In this work, we explore the vulnerability of data-driven SHM to some well-known existing adversarial attacks. As a result, we emphasize the importance of protecting against such attacks, particularly when ML models are used in sensitive tasks such as structural diagnosis and structural damage detection. Further, we discuss some mechanisms for avoiding these attacks while strengthening the robustness of models to adversarial examples. 
We propose adversarial training with circle loss, which optimizes feature distances to increase the distance between data points and decision boundaries.
The results of our experiments validate the effectiveness of our method in improving adversarial robustness by incorporating a feature distance constraint in the objective function, while conventional cross-entropy loss fails to impose. Finally, we encourage researchers to consider robustness to adversarial attacks when evaluating data-driven SHM models.




\section*{CRediT ~authorship ~contribution ~statement}
\textbf{Xiangli Yang}: Conceptualization, Validation, Formal analysis, Writing - Original draft preparation. \textbf{Xijie Deng}: Methodology, Investigation, Validation, Software, Writing - Original draft preparation. \textbf{Hanwei Zhang}: Conceptualization, Methodology, Validation, Writing - Original draft preparation. \textbf{Yang Zou}: Project administration, Review and editing. \textbf{Jianxi Yang}: Supervision, Funding acquisition.  

\section*{Declaration of Competing Interest}
The authors declare that they have no known competing financial interests or personal relationships that could have appeared to influence the work reported in this paper.

\section*{Acknowledgement}
This work was supported in part by the National Natural Science Foundation of China (Grant No. 62101081 and 62205039), the Science and Technology Research Program of Chongqing Municipal Education Commission (Grant No. KJQN202100747 and KJZD-M202300703), and the Graduate Research Innovation Project of Information science and Engineering School of Chongqing Jiaotong University (Grant No. 2023yjkc003).

\normalem
\bibliographystyle{elsarticle-num}

\bibliography{main}

\end{document}